# Supervised Deep Learning for Content-Aware Image Retargeting with Fourier Convolutions

MohammadHossein Givkashi[1], MohammadReza Naderi[1], Nader Karimi[1], Shahram Shirani[2], Shadrokh Samavi[1,2,3]

[1]Department of Electrical and Computer Engineering, Isfahan University of Technology, 84156-83111, Iran

[2]Department of Electrical and Computer Engineering, McMaster University, L8S 4L8, Canada

[3]Computer Science Department, Seattle University, Seattle 98122 USA

## Abstract

Image retargeting aims to alter the size of the image with attention to the contents. One of the main obstacles to training deep learning models for image retargeting is the need for a vast labeled dataset. Labeled datasets are unavailable for training deep learning models in the image retargeting tasks. As a result, we present a new supervised approach for training deep learning models. We use the original images as ground truth and create inputs for the model by resizing and cropping the original images. A second challenge is generating different image sizes in inference time. However, normal convolutional neural networks cannot generate images of different sizes than the input image. To address this issue, we introduced a new method for supervised learning. In our approach, a mask is generated to show the desired size and location of the object. Then the mask and the input image are fed to the network. Comparing image retargeting methods and our proposed method demonstrates the model's ability to produce high-quality retargeted images. Afterward, we compute the image quality assessment score for each output image based on different techniques and illustrate the effectiveness of our approach.

## 1. Introduction

The purpose of image retargeting is to alter the size of the images. In other words, image retargeting involves adapting images to different displays with a wide range of screen resolutions and aspect ratios without introducing distortion or artifacts. In traditional image resizing methods, pixels are treated similarly without considering the image's content. Recently, image retargeting methods generate images with a different aspect ratio while considering the image's content. Image retargeting is one of computer vision's most challenging and interesting tasks today. One of the best traditional image retargeting algorithms proposed in [1] is seam carving. Seam carving increases or decreases the image's size using an energy map generated from the input image. The algorithm identifies seams with low importance in the image, and the seams are iteratively added or removed to achieve the desired size. This method can be used for various images; however, some distortions appear in boundaries and objects. The authors in [2] introduced an approach to image retargeting that utilizes an adaptive 3D saliency model. This model incorporates 3D information into both seam carving and mesh warping strategies, resulting in more effective and accurate image understanding. By combining the adaptive 3D saliency with L1 norm of gradient in seam carving, the model generates an energy map for searching the least important seam. The authors suggested that analyzing big data can improve the accuracy of the 3D model, but hardware limitations require improving pretreatment on initial data from 3D cameras. They also proposed extending their framework into video retargeting processes due to its real-time performance advantages.

Deep learning has recently been applied to various computer vision tasks with excellent performance. As a result, researchers attempted to use deep learning for retargeting images. It is generally recognized that convolutional neural networks are the most advanced architecture for vision tasks. They can extract important features from images and use them for various purposes, including classification and segmentation. However, these architectures require a large dataset for training. Thus, a dataset is an essential factor when training deep learning models. However, a dataset containing ground truth for training deep learning models for image retargeting is unavailable.

Convolutional neural networks cannot generate an image with a different aspect ratio from the input image because of downsampling and upsampling in most common architectures. Hence, the output image's size is typically the same as the input image. Our ability to generate images of different sizes compared to the input image is crucial to the success of the image retargeting task. Among the advantages of deep learning models is that they can learn to extract features from a large number of images and that they require very little time during their inference phase to generate a desired output. Several one-shot learning approaches have been proposed to overcome the problem of the lack of paired images. However, one-shot learning is inefficient since each image must be trained separately. The authors in [3] developed a deep learning model that uses an input image to train, generating output images of the desired sizes after training each image separately. Because of this, their model requires a significant amount of time and resources to be trained, which is not useable in the real world. Furthermore, the model often fails to generate acceptable results, and distortion is evident.

A new approach for training convolutional neural networks for image retargeting has been proposed in this study to solve the problems mentioned above. Training deep learning models needs a paired dataset that contains input image and ground truth. To achieve this, we introduce a new method for creating input images and using the original images as ground truth. The reality is that images can be cropped and resized together or separately. As a result, we use an object detection model to crop the object in the input image and a resizing algorithm to change the image's aspect ratio.

Consequently, we use prepared input images for training the model and using original images as ground truth. So, the model can learn how to produce output like that. The main challenge for training a deep learning network for image retargeting is generating output images with different sizes and aspect ratios. Common deep-learning models can produce an output the same size as the input image. We introduce a new technique for generating images of various sizes to resolve this issue. As part of our approach, we use a mask that contains objects segmented from the original image. A zero-padded mask has been created to a specific size to increase the batch size during training. When we combine this approach with the changing size of the extracted object on a white background, we can obtain an output image in various sizes, which is the objective of image retargeting.

In summary, the main contributions of this work are:

- A novel supervised deep image retargeting approach is presented in this work, which uses the original image as ground truth and a mask that contains an object in the original image concatenated with an input image in the retargeting process.
- A new approach based on zero pad input images and masks with the object has been proposed to increase batch sizes in the training process and produce output images with a different aspect ratio, which is required for image retargeting.
- We present a network architecture that takes an image and a mask with an object as its inputs and generates a retargeted image in a single shot. Therefore, end-to-end training is possible, and the test time is fast. To the best of our knowledge is the first attempt to use a mask with an object and input image as inputs to the image retargeting application.
- We trained a deep learning model by creating a new dataset with images prepared with a unique technique.
- The proposed model can understand a shift in the object, which means the model can create a suitable output if the object shifts within the input mask.

The structure of this paper is as follows: In section 2, we will present some previous retargeting methods. Then, section 3 offers the proposed method. Finally, experiments are detailed in section 4, and the conclusion is presented in section 5.

## 2. Related Works

A review of previous studies related to this paper is presented in this section. Like most computer vision tasks, image retargeting research could also be classified as traditional and learning-based methods. We will discuss the proposed works in each category below.

### 2.1. Traditional methods

This subsection will review traditional methods based on content-aware cropping, seam carving-based, and shift-map methods.

#### 2.1.1. Content-aware cropping methods

Content-aware cropping approaches remove pixels from an image to create output with a desired size. This method crops the image with attention to the content, and selecting the appropriate cropping region is a key component of this approach. Thus, semantic details are extracted from the input image to select the most important content. The method proposed in [4] is based on important information about the image, and the greedy search is used to select the cropping window. Additionally, they used some information, such as facial information, in the image. As described in [5], an image attention model has been proposed to extract semantic information from both faces and text. Semantic information plays a significant role in the result of this method, and its extraction is the greatest challenge. Several optimization algorithms have been applied to image cropping. In [6], particle swarm optimization (PSO) has been applied to the automatic cropping problem to effectively determine the optimal solution and crop the image. In [7], the authors employed four feature extractors and a genetic algorithm to extract important regions within the image for the cropping task. Some works have used machine learning to identify an image's important regions. In [8], the authors proposed a method that used Support Vector Machine (SVM) to extract a saliency map and required information. Therefore, they attempted to select the best cropping window. In [9], a classification and regression tree classifier has been applied to image cropping. Images are classified into semantic types, such as landscapes and close-ups. The method creates output results based on semantics and visual information. In [10], an important region of the image was considered to be the region of attention. This region was semi-automatically identified based on the user's gaze when viewing an image. Hence, the cropping algorithm can perform better and preserve important contents. The authors in [11] proposed a method that can be used to search for the cropping window and determine which region contains the greatest saliency density without understanding the shape or size of the salient objects in advance.

A content-aware image cropping method also changes the image size without structural change. However, there is a strong correlation between the performance of these methods, the number of objects, and the important regions. Therefore, they cannot perform well with one cropping window [12].

**2.1.2. Seam-carving-based methods**

The seam carving algorithm [1] is one of the traditional critical algorithms for content-aware image retargeting. The seam carving technique involves calculating an energy map from the input image and adding or removing pixels with similar energies iteratively to reach the desired size. This process can be used to resize images vertically or horizontally. The authors in [13] presented a new technique for resizing images that considers gradient, saliency, and depth information to create an energy map. The method determines the importance of each map using an algorithm that assigns an importance coefficient to each one. The energy map is then used to find a threshold on seam energy to switch from seam carving method to scaling method, which helps maintain the geometrical structure of image objects. The results demonstrate that this approach outperforms previous methods in terms of preserving important details and producing better global appearance, making it a promising strategy for image retargeting. The authors in [14] proposed a method for finding seams efficiently for the seam carving method and reducing visual artifacts in the output image. Therefore, the optimal seam is one where the minimum energy is introduced into the image after removing it. A method for interactive image retargeting has been proposed in [15]. This framework uses a structure-aware energy field instead of the conventional energy field to extend seam carving frameworks. As a result, essential aspects of the image, such as shape boundaries, can be preserved. Using the genetic algorithm (GA) with the user interaction, the authors designed a graphical user interface (GUI) for better optimization. The authors in [16] proposed a method for sports image resizing. In their approach, important image regions are segmented to recognize semantic edges based on semantic information extracted from the image. The seam carving method uses semantic information to create the output image. The authors in [17] introduced a new method to tackle the issue of distortions that occur during image retargeting using the seam-carving algorithm. The proposed continuous model employs distortion detection and mean field approximation to produce a better retargeting outcome. The authors suggested that future work should focus on identifying the optimal ratios of multiple features. The authors in [18] used an importance map as a combination of gradient, saliency, skin, and canny edge maps. Additionally, they used an organized approach that combined seam carving and scaling to balance information loss and image stretching. There has been used a nonparametric semantic segmentation method in [19], which is used for content-aware image retargeting aimed at identifying the most important map within the input image. In contrast to the parametric semantic segmentation method, there is no need for a global learning model with the nonparametric method. Thus, a combination of seam carving and scaling produces retargeted output. The authors in [20] presented an enhanced approach to seam carving-based image resizing that considers the shadows present in the image as important information. The quality of the importance map extracted from the image is crucial for the performance of seam carving algorithms, but previous methods have not taken into account the significance of shadows. The proposed method preserves shadows while resizing images, leading to a better and faster understanding of image content.

### 2.1.3. Shift-map methods

Shift-map is used to create patches from the input image. This method uses discontinuous displacement or removal of similar patterns in place of scaling and stretching. The graph cuts method is used to distribute the content throughout the image evenly. This method is acceptable but less effective if the image contains large or different objects. Authors of [21] propose resizing an image by determining regions based on shift maps. An optimal "shift map" can be conceptualized as graph labeling, where each graph node represents a pixel of the output image. Their method removes or shifts the same patterns in the input image to achieve the desired size. In addition, their approach cannot perform well when many different objects are in the input image. The authors in [22] proposed a new approach to video retargeting called Hybrid Shift Map, which maximizes the spatial-temporal naturality between source and target videos without requiring motion analysis. The method employs a multi-resolution framework that enhances computational efficiency while producing satisfactory retargeting results. The 3D shift map is used to obtain the spatial-temporal naturality measure. An incremental 2D shift map is applied to refine every seam in each frame with temporal consistency, resulting in improved retargeting outcomes. The authors in [23] presented a novel approach to retargeting stereo images using the shift-map image editing framework. The proposed method incorporates a stereo correspondence constraint to preserve pixel correspondences between the stereo pair and maintain the scene's structure. Experimental results demonstrate the effectiveness of the approach. Shift-map image editing can be used for retargeting, inpainting, reshuffling, and image composition.

## 2.2. Learning-Based Retargeting Methods

In this part, we will review some learning-based retargeting methods based on deep learning and one-shot learning methods that use the input image to create the desired output.

### 2.2.1. Deep learning-based methods

Supervised deep learning models require a large dataset consisting of paired images. However, some researchers have attempted to use deep learning models as a component of their proposed approach for using deep learning in image retargeting tasks.

The authors in [24] used image context and semantic segmentation to determine saliency from the input image. A convolutional neural network has created a saliency map. The original image and saliency map were used as input to an adapted version of the pixel fusion approach presented in [25]. The authors in [26] proposed Cycle-IR, which contains a cycle in the architecture and uses an attention mechanism. However, this method cannot handle large or lacking contrast images when their visually important regions are large or scattered. The Deep Image Retargeting (DeepIR) method [27] creates the semantic structure of the input image with a convolutional neural network. The search algorithm employs the semantic structure to create the desired image.

DeepIR can preserve some important content in the image, but it can also lead to the over-compression of essential parts of the image or the saving of pointless areas. A large-scale image retargeting dataset for supervised learning does not exist, but some researchers attempted to use supervised learning differently. A deep learning approach for image retargeting is introduced in [28]. In this case, the method is performed by a shift map consisting of a pixel-by-pixel mapping between the source and target grids. The model learns attention maps to preserve essential regions in the retargeting process, and their architecture contains encoder and decoder parts. When images are encoded, high-level information is extracted from them, and the decoder generates a map of attention. The model generates the desired output based on the regions the decoder considers essential to the decoding process.

The authors in [29] proposed a reinforcement learning method for fast image retargeting. So, they developed an agent with a reward function that can determine the best action between seam carving, scaling, and cropping at different stages of the production process to create the final output. One of the comparison methods used is MULTIOP [30], which selects different operators to make the final output image. Consequently, their reinforcement learning approach can achieve a faster processing time than MULTIOP [30]. The authors in [31] proposed an image retargeting method based on reinforcement learning. Retargeting is a series of operations performed on an image to achieve a desired result, such as seam carving, scaling, and cropping. In the reinforcement learning model, these three operators were considered actions and rewarded based on semantic and aesthetic measurements. A limitation of their model is that it cannot generate images in different scale factors.

### 2.2.2. One-shot learning methods

Several studies have attempted to use a one-shot learning method for training deep learning models using the input or a single image to create the output using deep learning techniques.

Generative adversarial networks (GANs) have been used in different computer vision tasks to generate high-quality images. GANs perform effectively for unsupervised learning [32]. The authors in [3] proposed a model based on a generative adversarial network for image retargeting. The model uses an input image in the training process to generate desired output size. As a result, the model must be trained for each input image separately and need remarkable time in the training process. After learning the internal distribution of an image, the model can retarget that image at different sizes. Also, by cropping and adding noise to the image, their model learns the internal distribution of the image, which ensures it does not fully understand the scene. It is important to note that the model duplicates every object in the image when it is resized to a larger size. Consequently, the output in most cases is undesirable, and distortion is evident.

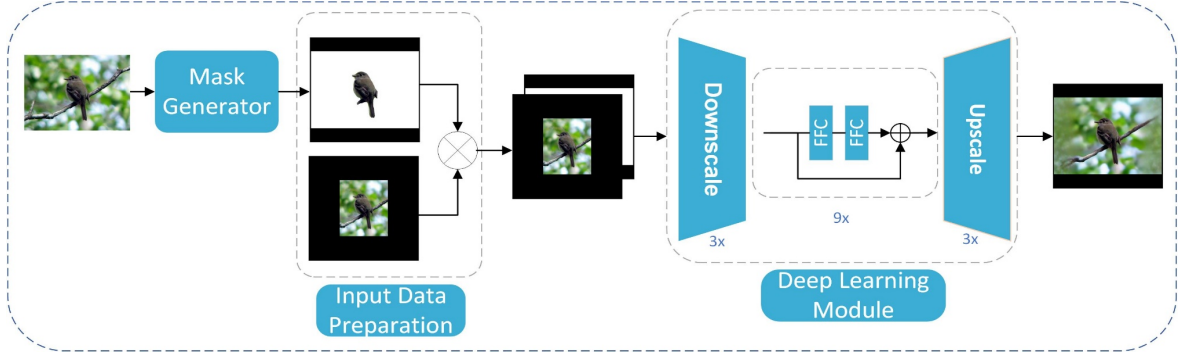

**Fig. 1.** Block diagram of the proposed method.

In [33], adversarial learning is a method that can be used for various tasks, including super-resolution, denoising super-resolution, and image retargeting. According to their work, the one-shot learning method proposed, similar to [3], retargets the input image to different sizes by learning the internal distribution of the image. In their retargeting approach, they suffer from the same problem as [3], which repeats objects in large image sizes. The authors in [34] proposed an object-aware image retargeting method. First, the objects are extracted using a binary segmentation mask, and then a pre-trained image inpainting model is used to fill in the gaps. Then, the seam carving algorithm [1] is used to create the background image of the desired size. The super-resolved foreground image was used to replace the background image to achieve an aesthetically pleasing result. Finally, the particle swarm optimization algorithm [35] was applied to find the best place and size of the foreground image

The authors in [36] proposed a one-shot learning method that could be applied to different tasks, such as image generation, conditional inpainting, image retargeting, etc. For image manipulation, methods based on single-image GANs require significant training time. Furthermore, there is a need to train separately for each image and task, and they may suffer from visual artifacts. Therefore, they developed a patch-based method called GPNN for creating the desired output based on the input image in a limited time. As part of their method, patches are made from the input image, and the patch nearest neighbor approach is used to create the output image. It is important to note that the output of their method depends on the input image, and in some cases, distortions and artifacts may be evident. The authors in [37] used the same approach as [36] for image generation tasks. A new algorithm is developed that minimizes the distance between patch distributions in two images explicitly and efficiently using Sliced Wasserstein Distance. Various tasks were carried out using the proposed method, including style transfer, reshuffling, and image retargeting. In their approach, called GPDM, patches from the input image are copied and aggregated to generate the output image based on the closest patches to the current estimated image.

## 3. Proposed Method

Image retargeting changes the size of an image while maintaining visually interesting areas in the input image without significant distortion. In this work, we propose a method based on deep learning for image retargeting. This method introduces a new technique for creating output images with different sizes than input images using convolutional neural networks and creating a paired dataset for training deep learning models. An input mask with an object determines the size of the output image and where the object will be located. The model requires a particular input format incorporating the mask containing objects and the input image. Each of them comprises three channels, implying that when stacked together, these two images create six channels for the model to process. So, the retargeted image is generated as the output of the model. The proposed architecture is illustrated in Fig. 1. There are three components in the proposed method: a mask generator, input data preparation, and a deep learning module. First, we will discuss the mask generator and how to prepare input data in subsections 3.1 and 3.2. Then, in subsection 3.3, the architecture of the deep learning model will be discussed. The loss functions for training the model will be discussed in the final subsection, 3.4.

### 3.1. Mask Generator

For the image retargeting task, output images are generated with different sizes compared to the input image. However, convolutional neural networks are sensitive to the size of the input image during the training process. Because of this, using one image in each iteration of the training process is the common way to train convolutional neural network models with different input image sizes. Consequently, we propose a novel approach for training convolutional neural networks with images of varying sizes. First, we determine the largest width and height of the images in the dataset to set the input size for training the model. In our approach, we use 512 pixels for width and height. Next, we use a pre-trained image segmentation model named Deeplabv3 [38], which extracts objects from input images and places them on a white background the same size as the original image. Then, the input mask will be zero-padded to arrive at the size we select for the training process.

### 3.2. Preparation of training data

Deep learning models typically require large image datasets for training. No paired dataset is available to train deep-learning models for image retargeting. Our method uses original images from the dataset as ground truth to train a deep learning model. Original images are resized and cropped simultaneously to create input images for retargeting training. When cropping an image, it is necessary to save the objects within the cropped image. We utilize Yolo's pre-trained [39] object detection model to accomplish this goal. In the Yolo model, objects are extracted from an image. With the object extracted, the surrounding part of the image can be cropped or resized within a certain range simultaneously.

### 3.3. Deep learning model

The most commonly used architecture in computer vision is the convolutional neural network. They effectively extract critical features from the input image and are considered state-of-the-art in most vision applications. It is important to note that the convolutional neural network analyzes the image in a localized manner since the size of the kernel plays the most significant role in feature extraction. Therefore, we use Fast Fourier Convolution (FFC) [40], which was recently proposed.

FFC utilizes global context in the early layer and is based on a channel-wise Fast Fourier Transform [41], with a receptive field encompassing the entire image. In the early layers, FFCs assist the generator in using a global context. For image retargeting, it is important to change the size of the image while paying attention to its content. Our model is similar to a ResNet [42] architecture with 3 downsampling blocks, 9 residual blocks, and 3 upsampling blocks. We use FFC in residual blocks, which help the model increase the receptive field and an NLayerDiscriminator [43], to determine whether the generated image looks real or fake. The model receives an input comprising six channels, namely the input image and the mask with the object, where each has three channels concatenated together. The downsampling and upsampling blocks facilitate the extraction of features from the input and subsequently employ them to generate the output image. We use Adam [44] optimizer, with learning rates of 0.001 for the generator and 0.0001 for the discriminator network.

### 3.4. Loss functions

Image retargeting is a visually important task in computer vision which needs different loss functions to be trained well. This part will discuss different parts of the loss function used in the training process.

**3.4.1. High receptive field perceptual loss**

The perceptual loss [45] evaluates the distance between features extracted from the predicted and target images using the pre-trained model $\phi(\cdot)$. In this Equation, $x$ and $\hat{x}$ present predicted image, ground truth. We use high receptive field perceptual loss (HRFPL), which uses the base network with a fast growth of receptive field introduced in [46]. The formula of HRFPL shows below:

$$\mathcal{L}_{HRFPL}(x, \hat{x}) = \mathcal{M}([\phi_{HRF}(x) - \phi_{HRF}(\hat{x})]^2)$$

where $\mathcal{M}$ is the sequential two-stage mean operation. The pre-trained network for computing the perceptual loss is important because using a segmentation model as a backbone for perceptual loss may help the model focus on high-level information. On the contrary, classification models are known to focus more on textures.

**3.4.2. Adversarial loss**

We use an adversarial loss to ensure that the model generates naturally looking local details. We use a discriminator [43], which sees the image as patches, that determines whether each patch is real or fake. In Equations below, $x, \hat{x}, D, G, \varepsilon, \theta$, and $sg$, present predicted image, ground truth, discriminator, generator, parameters of the discriminator, parameters of the generator, and stop gradient to certain parameters, respectively. The formula of the adversarial loss illustrates below:

$$\mathcal{L}_D = -\mathrm{E}_x[\log D_\varepsilon(x)] - \mathrm{E}_x[\log(1 - D_\varepsilon(\hat{x}))]$$

$$\mathcal{L}_G = -\mathrm{E}_x[\log D_\varepsilon(\hat{x})]$$

$$\mathcal{L}_{Adv} = sg_\theta(\mathcal{L}_D) + sg_\varepsilon(\mathcal{L}_G) \rightarrow \min_{\theta,\varepsilon}$$

### 3.4.3. The final loss function

In the final loss function, we use $\mathcal{R}_1 = \mathbb{E}_x \left\| \nabla \mathcal{D}_\xi(x) \right\|^2$ gradient penalty [47–49] and feature matching loss, is the perceptual loss on the features of discriminator network $\mathcal{L}_{DiscPL}$ [50].

The final loss function for our image retargeting network is the weighted sum of introduced losses. Each part is responsible for attention to the different features in the output image.

$$\mathcal{L}_{final} = \kappa \mathcal{L}_{Adv} + \alpha \mathcal{L}_{HRFPL} + \beta \mathcal{L}_{DicsPL} + \gamma \mathcal{R}_1$$

## 4. Experiments

In this section, we will explain the training process in subsection 4.1 and the results of our proposed method in comparison with other retargeting methods in subsection 4.2.

### 4.1. Training

We use CUB [51] to train the proposed model, a dataset that contains 11,788 bird images of 200 categories. The experimentation pipeline is implemented using PyTorch [52] library. The input images are zero-padded to the 512 x 512 for training the model. It takes around 2 ~ 3 days for 40 epochs with an Nvidia GeForce RTX 3090 GPU with batch size 6.

### 4.2. Results

In this part, we review our results. Fig. 2 shows several examples of retargeting output from different methods. Some images used for comparision with other methods and prepared similarly to the training process using resizing and cropping techniques. The output images are the same size as the ground truth, and we can see that the outputs of our proposed method have high quality compared with seam carving [1], GPNN [36] and GPDM [37] which they have ability to create retarget image in many different sizes. The seam carving [1] method destroyed the boundaries and made artifacts in the output image. The GPNN [36] and GPDM [37] methods tried to repeat pixels in the image to achieve the desired size. As we can see, distortion and artifacts are evident in the output images.

One of the useful features of our proposed mothed is that the model can understand the shift in the input mask with an object to generate an output image with attention to the position in the input mask image. Therefore, the users can select the position that they want to replace the object in there and retargets the image in the desired size. Fig. 3 illustrates how the model can understand the shift and generate output. For two images, we can see the results. (Fig. 3-(a),(b)) and (Fig. 3-(c),(d)) show the input image and mask with an object which shifted differently, and (Fig. 3-(c)) and (Fig. 3-(f)) show the generated output from our model. The dashed line has been drawn for better shift visualization in masks with object images.

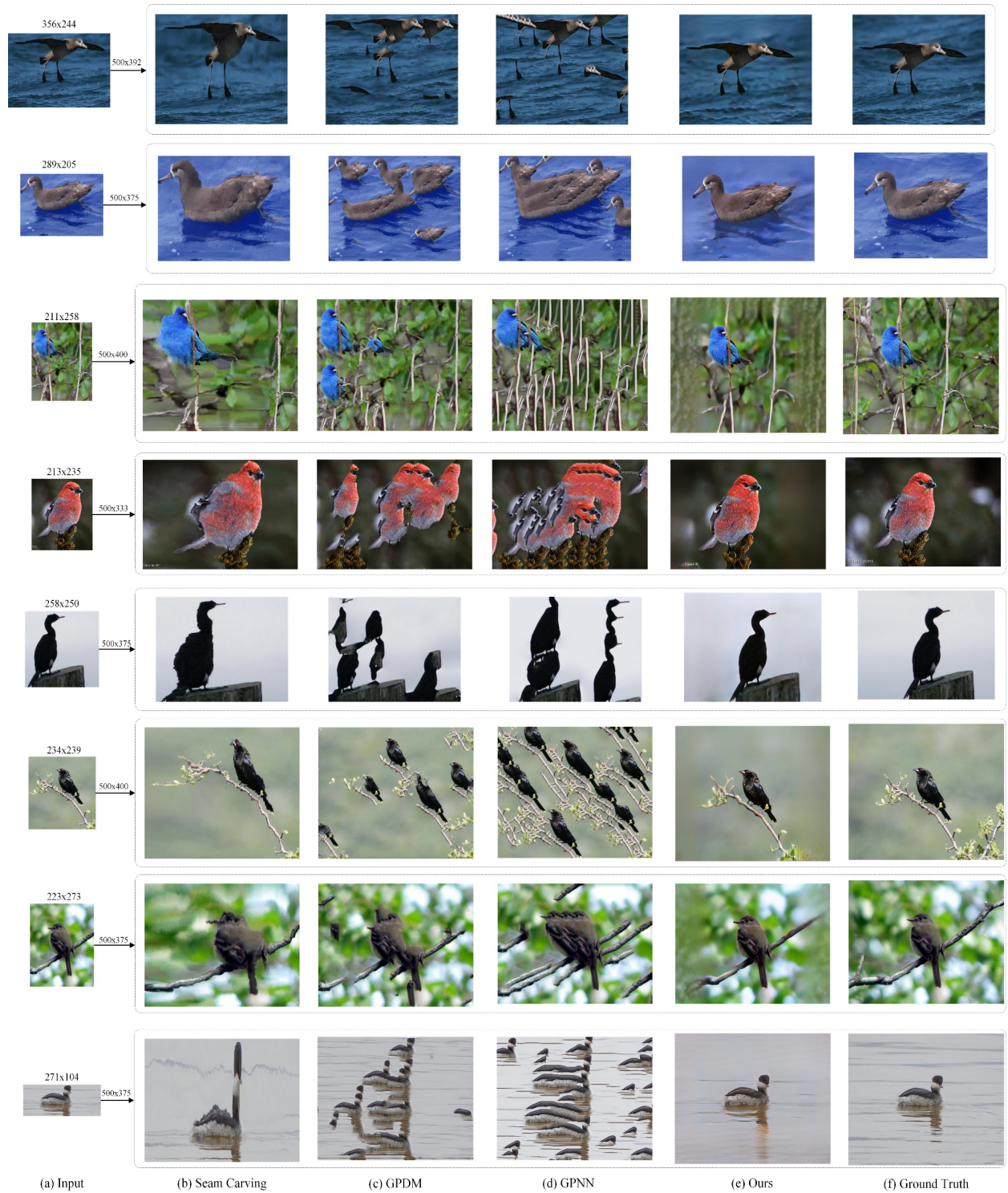

**Fig. 2.** Comparing results of our proposed method with Seam Carving [1], GPNN [36] and GPDM [37].

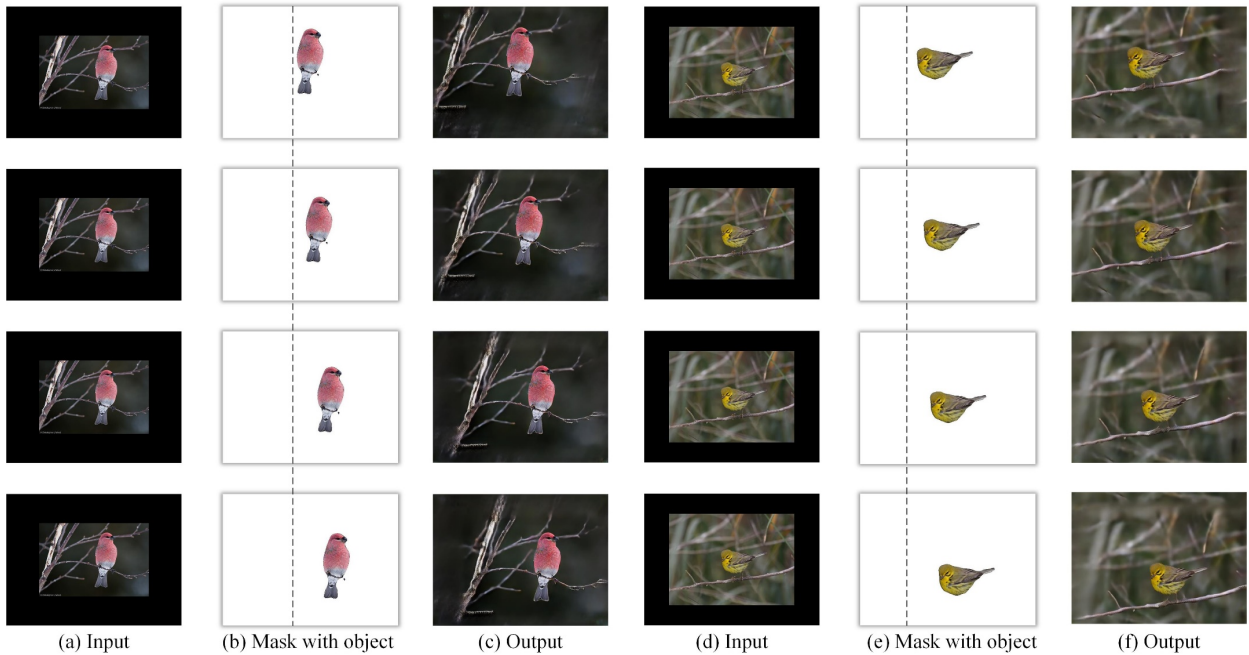

(a) Input    (b) Mask with object    (c) Output    (d) Input    (e) Mask with object    (f) Output

**Fig.3.** The results of shifted objects. (a),(d) Input images,(b),(e) Mask with object and (c),(f) Output images.

Fig. 4 and Fig. 5 illustrate the output images generated with random sizes to show the model's capability for creating images in different sizes. The input images were made the same as the training process, using resizing and cropping simultaneously. In Fig.4, column one, the input images have been used for different methods to retarget the input images to the random size. As we can see for row one, the output of the seam carving algorithm is not acceptable because the object can not preserve in the suitable shape as the input image. On the other hand, the outputs from GPDM and GPPN methods show that these methods just tried to repeat pixels to achieve desired sizes. Therefore, the output of our proposed method preserves objects well and retargets images to the desired size. One way to measure image quality is by using an image quality assessment network. The image quality assessment model gives the score for the input image; a higher score shows better image quality. We used a pre-trained image quality assessment model [53] to measure the quality of the images from different methods. In Fig. 4 and Fig. 5, the score for each image is represented. As we can see, the score of the output images from our proposed method is higher in most cases. As a point of clarification, the model for assessing image quality has not seen the images like the output of GPNN during the training process. Thus, it cannot perform well for this type of image in which the object is repeated. The code for the proposed method is available at https://github.com/givkashi/CAIR.

## 5. Conclusion

In this paper, we proposed a new supervised learning approach for training deep learning models in image retargeting tasks. The model is capable of producing visually pleasing retargeted images. Our model generates a retargeted image with attention to the input mask that contains an object in a particular location. The network architecture consists of the encoder-decoder part using Fast Fourier Convolution, which helps the model to have a high receptive field to generate better results and achieve a global understanding of the images. The model can create images in different sizes by changing the input mask size with the object. Also, we introduced a new approach for training deep learning models with specific dimensions and getting output with different sizes. The perceptual and adversarial losses helped the model preserve essential objects and reduce visual artifacts such as boundary distortion. The experimental results illustrate that our proposed method is superior to other image retargeting approaches and shows more visually pleasing qualitative results.

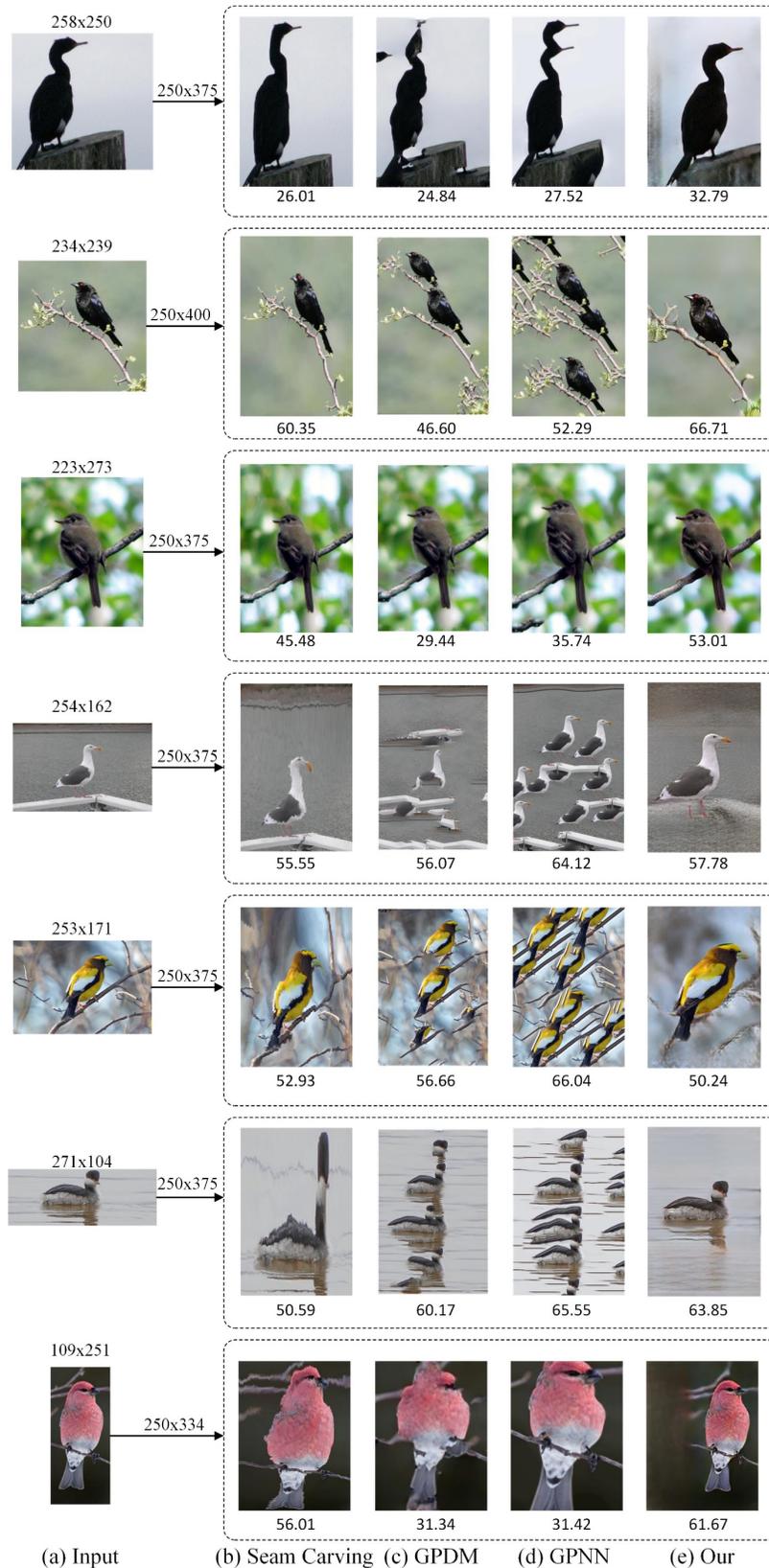

**Fig.4.** Visual comparison of our method with representative retargeting approaches. The images are retargeted with random widths and heights.

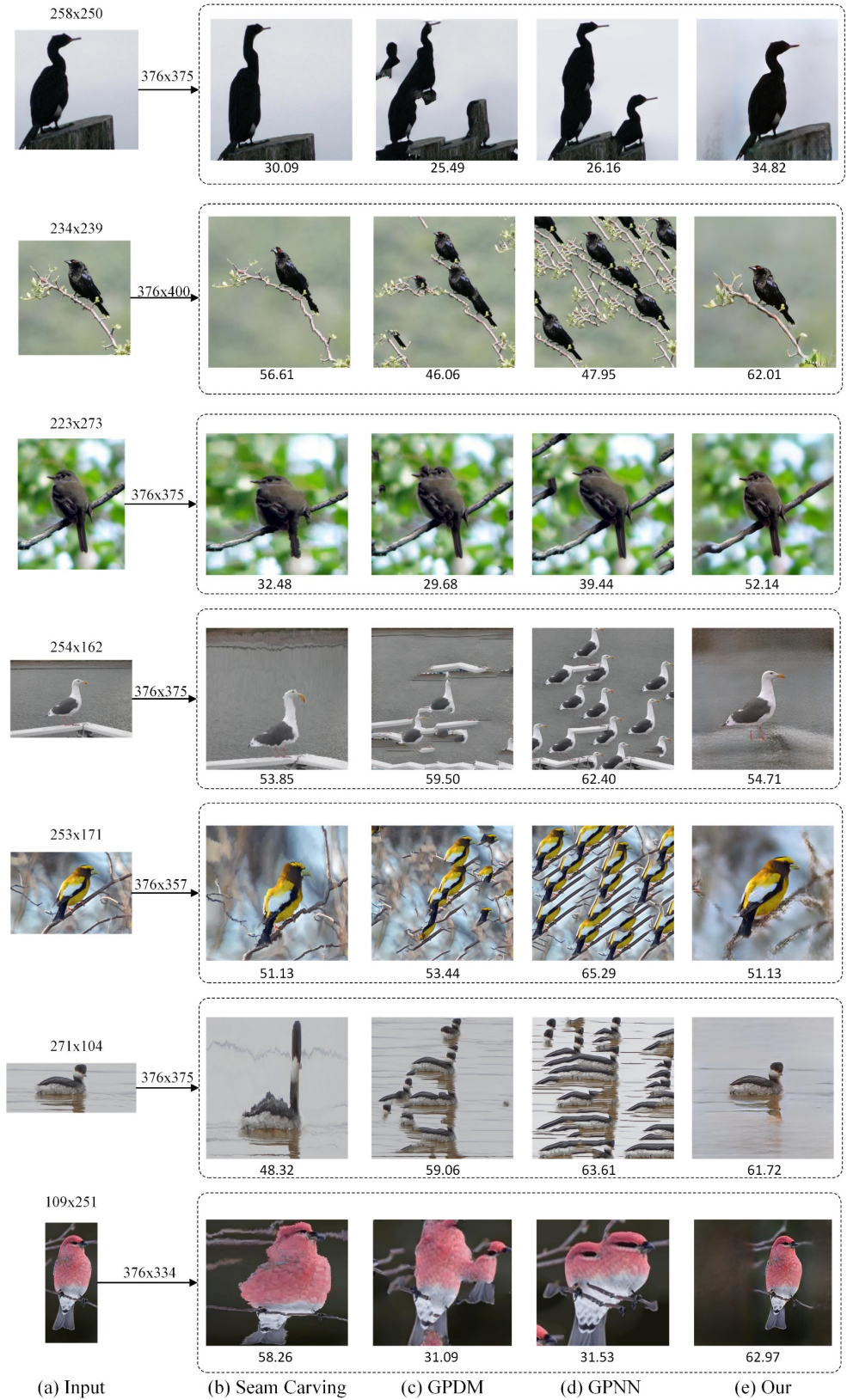

**Fig.5.** Visual comparison of our method with representative retargeting approaches. The images are retargeted randomly in width and height.